\pdfoutput=1
\documentclass[11pt]{article}
\usepackage{amsmath}
\usepackage[preprint]{acl}
\usepackage{caption}
\usepackage{tabularx}
\usepackage{times}
\usepackage{latexsym}
\usepackage{subcaption}
\usepackage{arydshln}
\usepackage{microtype}
\usepackage{multirow}
\usepackage{soul}
\usepackage{booktabs}

\usepackage{inconsolata}

\usepackage{graphicx}
\usepackage[T1]{fontenc}

\usepackage[utf8]{inputenc}

\usepackage{microtype}
\usepackage{pifont}
\usepackage{adjustbox}
\usepackage{float}
\usepackage{inconsolata}
\usepackage{tikz} 
\usepackage{enumitem}

\definecolor{main}{HTML}{5989cf}    
\definecolor{sub}{HTML}{cde4ff}     
\usepackage[many]{tcolorbox}[enhanced,breakable]
\newtcolorbox{boxM}{
    use color stack,
    enhanced,
    breakable,
    fontupper = \color{white},
    fontlower = \color{white},
    rounded corners,
    arc = 6pt,
    colback = main!80, 
    colframe = main, 
    boxrule = 0pt, 
    bottomrule = 4.5pt,
    fuzzy shadow = {0pt}{-3pt}{-0.5pt}{0.5pt}{black!35}
}

\newtcolorbox{boxE}{
    enhanced, 
    breakable,
    boxrule = 0pt, 
    borderline = {0.75pt}{0pt}{main}, 
    borderline = {0.75pt}{2pt}{sub} 
}

\newtcolorbox{boxK}{
    enhanced, 
    breakable,
    sharpish corners, 
    boxrule = 0pt,
    toprule = 4.5pt, 
    enhanced,
    fuzzy shadow = {0pt}{-2pt}{-0.5pt}{0.5pt}{black!35} 
}
\usepackage{colortbl}
\title{A Closer Look at Tool-based Logical Reasoning with LLMs: \\The Choice of Tool Matters}

 \author{Long Hei Matthew Lam\ \ \ \ \ \ \ Ramya Keerthy Thatikonda\ \ \ \ \ \ \ Ehsan Shareghi\\
 DSAI, Monash University\\
\texttt{llam0013@student.monash.edu.au}\ \ \ \ \ \ \ \texttt{Ramya.Thatikonda1@monash.edu}\\
\texttt{Ehsan.Shareghi@monash.edu}}

\begin{document}
{\makeatletter\acl@finalcopytrue
  \maketitle
}
\begin{abstract}
The emergence of Large Language Models~(LLMs) has demonstrated promising progress in solving logical reasoning tasks effectively. Several recent approaches have proposed to change the role of the LLM from the reasoner into a translator between natural language statements and symbolic representations  which are then sent to external symbolic solvers to resolve. This paradigm has established the current state-of-the-art result in logical reasoning (i.e., deductive reasoning). However, it remains unclear whether the variance in performance of these approaches stems from the methodologies employed or the specific symbolic solvers utilized. There is a lack of consistent comparison between symbolic solvers and how they influence the overall reported performance. This is important, as each symbolic solver also has its own input symbolic language, presenting varying degrees of challenge in the translation process. To address this gap, we perform experiments on 3 deductive reasoning benchmarks with LLMs augmented with widely used symbolic solvers: Z3, Pyke, and Prover9. The tool-executable rates of symbolic translation generated by different LLMs exhibit a near 50\% performance variation. This highlights a significant difference in performance rooted in very basic choices of tools. The almost linear correlation between the executable rate of translations and the accuracy of the outcomes from Prover9 highlight a strong alignment between LLMs ability to translate into Prover9 symbolic language, and the correctness of those translations.\footnote{Code and data are publicly available at \url{https://github.com/Mattylam/Logic_Symbolic_Solvers_Experiment}.}



\end{abstract}

\section{Introduction}
%
The recent state-of-the-art approaches to logical reasoning have combined Large Language Models~(LLMs) with external symbolic mechanisms \cite{Nye, logicLM,satlm,ProglM, FaithfulCof}. This approach leverages LLMs' remarkable proficiency in translating natural language into symbolic representation such as First Order Logic~(FOL) or symbolic solvers' specified language~(e.g., Pyke, Z3) \cite{NL-FOL}, and the symbolic solver’s ability to execute these translations through a 
fully deterministic proof process~\cite{Expert-System}. These existing published methods try a variety of tools and tool-specific formalism.
Table \ref{Tool Summary Table} summarises various tools used in recent state-of-the-art studies. This variability of tools makes it impossible to have a fair understanding of each approach. There is currently a lack of consistent comparison that will allow others to understand better where this performance gain stems from. 
%

In this paper, we take 3 widely used tools: Z3 \cite{Z3}, Pyke \cite{Pyke}, and Prover9 \cite{mccune2005release} 
and analyse the difficulty LLMs face for translating natural language into their desired input format, and the internal capability of these tools at solving certain satisfiability tasks. We select GPT4o, GPT-3.5-Turbo \cite{gpt3}, Gemini-1.0-Pro \cite{Gemini} and Cohere Command R Plus, as representatives of the most capable family of LLMs, along with 3 widely used deductive reasoning benchmarks ProofWriter \cite{ProofWriter}, FOLIO \cite{FOLIO}, and ProntoQA \cite{ProntoQA}. We conduct a fair side-by-side comparison of tools by trying various number of identical prompts, demonstration shots, and minimal adjustment for each solver. 

Our findings indicate that LLMs find it easier to translate for Prover9, followed by Z3, and lastly Pyke. 
Although Prover9 can solve more questions accurately, Prover9 demonstrates a lower discrepancy between execution rate and overall accuracy. This means that Prover9 is more likely to solve a question given the right syntax and format produced by LLMs. Overall, Z3 and Prover9 are all competitive options, Pyke's performance is significantly inferior and only comparable to the other tools in solving PrOntoQA. Our experiments across 3 benchmarks (based on the accuracy of outputs) highlight an up-to 50\% of performance variation for each LLM under different tools, and well as the performance change for each tool under different LLMs.

\begin{table}[t]
\begin{adjustbox}{width=\linewidth,center}
\begin{tabular}{l|l|l|l}
\hline
\textbf{Solver}             & \textbf{Dataset}                                                                                  & \textbf{Papers}                                                           & \textbf{Problem}                                                           \\ \hline
Z3                 & \begin{tabular}[c]{@{}l@{}}AR-LSAT \cite{AR_LSAT},\\ ProntoQA \cite{ProntoQA},\\ ProofWriter \cite{ProofWriter},\\ BoardgameQA \cite{BoardGameQA}\end{tabular} & \begin{tabular}[c]{@{}l@{}}LogicLM,\\ SatLM\end{tabular}          & \begin{tabular}[c]{@{}l@{}}Analytical,\\ Deductive,\\ FOL\end{tabular} \\ \hline
Pyke               & \begin{tabular}[c]{@{}l@{}}ProntoQA \cite{ProntoQA},\\ ProofWriter \cite{ProofWriter}\end{tabular}                           & \begin{tabular}[c]{@{}l@{}}LogicLM,\\ Logical Solver \end{tabular} & \begin{tabular}[c]{@{}l@{}}Deductive,\\ FOL\end{tabular}               \\ \hline
Prover9            & FOLIO \cite{FOLIO}                                                                                     & \begin{tabular}[c]{@{}l@{}}LogicLM,\\ LINC \end{tabular}                                                            & \begin{tabular}[c]{@{}l@{}}Deductive,\\ FOL\end{tabular}             \\ \hline
\end{tabular}
\end{adjustbox}
\caption{A summary of the symbolic solvers and the datasets it has solved in different studies: LogicLM \cite{logicLM}, LINC \cite{linc}, Logical Solver \cite{logical_solver}, and SatLM \cite{satlm}.}
\label{Tool Summary Table}
\end{table}

\section{Tools \& Logical Reasoning with LLMs}

\begin{figure*}[ht!]
  \centering
  \includegraphics[width=\textwidth]{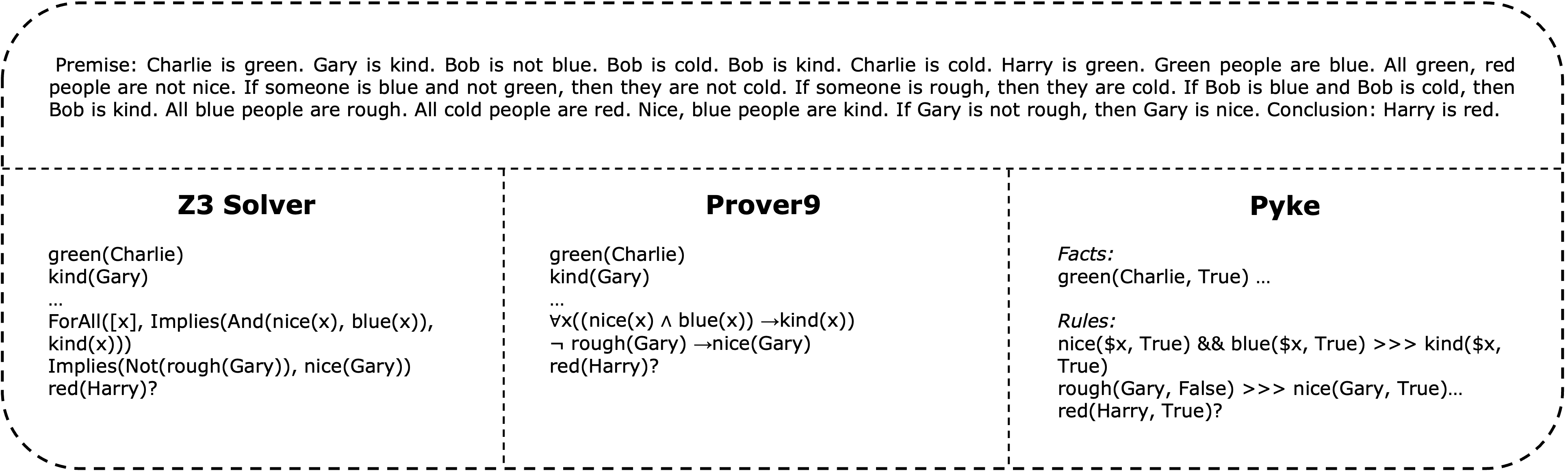}
  \caption{Overview of syntax used for different Theorem Provers: Z3 and Prover9 adhere to the traditional first-order logic (FOL) format, while Pyke adopts a simplified formula approach, distinguishing premises into rules and facts}
  \label{fig:proversyntax}
\end{figure*}

The tool-based approaches to logical reasoning combine LLMs with external symbolic solvers. This synergy harnesses the capability of LLMs to convert diverse natural language statements into logical symbolic formalism. While being less flexible compared with free-form reasoning methods, such as Chain-of-Thought~\cite{CoF}, the tool-based approach, given a \emph{correct formal translation}, has important advantages: logical coherence during the reasoning~(i.e., unlike LLMs, theorem provers cannot make reasoning shortcuts or hallucinate) is guaranteed, while the internal proof trace of the theorem provers offers a transparent and verifiable reasoning chain. 

\subsection{Logical Solvers}
Automated theorem provers (ATPs) and Satisfiability Modulo Theories (SMT) solvers are tools equipped with built-in functions designed to assist in logical reasoning tasks. These solvers can vary in syntax, proof search strategies, theorem automation, and complexity. ATPs efficiently resolve first order logic problems without external interaction. SMT solvers closely resemble ATPs in solving first-order formulae but add complexity by handling theories such as equality, arrays, and bit-vectors. Logical solvers, specifically Z3, Prover9, and Pyke, are used for logical reasoning tasks with LLMs due to their ease of use in a Python environment \cite{logicLM,satlm}. We study the logical solvers based on their ability to handle first-order logic and explore the crucial differences in external syntax and internal theories of these tools. In this context, we define the task as follows: given a set of premises \(P \in \{P_1, P_2, \ldots, P_n\}\), the objective is to determine whether the conclusion \(C\) logically follows from these premises. The translation syntax for each tool is presented in~Figure\ref{fig:proversyntax}.
\paragraph{Z3 Prover} developed by Microsoft, is an SMT solver designed to determine the satisfiability of given constraints \cite{Z3}. Z3 encompasses a diverse array of functionalities, including equality reasoning, arithmetic operations, handling arrays, and incorporating quantifiers. It supports multiple programming languages and mathematical operators, making it a versatile tool for a wide range of research applications. Z3 utlizes the DPLL algorithm for satisfiability resolution, where constrains are converted to conjunctive normal form (CNF). The solver then searches for a solution through backtracking, continuing until it finds a combination of truth values that satisfies the conditions. In deductive logical reasoning, the tool can check if the conclusion \(C\) renders the assertions \(P\) satisfiable. Z3 requires an explicit specification of data types of variables, functions, and their attributes, which are typically Boolean for deductive reasoning. Due to its flexible operations, Z3 has been applied to tasks beyond logical verification, as shown in Table \ref{Tool Summary Table}. Additionally, the simplicity of these tasks enables the translation format of Z3 to resemble programming languages, as demonstrated in Appendix \ref{Prompts}. 

\paragraph{Prover9} is an automated theorem prover for first-order and equational logic, based on resolution techniques \cite{mccune2005release}. This tool accepts first-order logic statements and applies logical transformations such as CNF conversion, quantifier operations, and skolemization to produce simplified clauses. The inference process involves iterating over given clauses to generate new clauses in a non-redundant manner by categorizing the clauses into usable and non-usable forms. For deductive reasoning task, the premises \(P\) produce new premise, i.e., \(\{P_1, P_2\} \implies \{P_{12}\}\), for various combinations. These derived premises \(P_{xy}\) are retained if they are relevant to the conclusion, and discarded otherwise. The inference is based on all the stored premises once all combinations have been exhausted. Although the logical transformations allow flexible input, Prover9 is sensitive to special characters and spaces, which require careful handling. Compared to Z3 or other ATPs, Prover9 cannot solve a variety of mathematical problems, thus limiting its applicability to certain fields of logic \cite{mccune2003otter}. In Python, Prover9 is accessible through the NLTK logic library. 


\paragraph{Pyke} short for Python Knowledge Engine, is a solver used for building and executing rule-based expert systems \cite{Pyke}. Although pyke is used for optimizing software development, \citet{logicLM} demonstrated its application in solving a first-order logic problem. Given a logical inference task, Pyke establishes a knowledge base and incorporates known facts (\texttt{fact.kfb}) and rules (\texttt{rule.krb}) from the input, i.e., \(P \xrightarrow{} (P_{\text{facts}}, P_{\text{rules}})\). The conclusion is parsed as a rule that is propagated through the knowledge base until it reaches a resolution. The predicates in the first order logic are treated as facts and are connected to form rules. Given its limited syntax, Pyke supports simple connectives such as `and', `or', and `implies'. The free variables (e.g., \(\$x\)) are generally considered to be universal quantifiers, thus restricting the use of existential quantifiers. Due to these limitations, Pyke may not adequately handle complex tasks involving first-order logic, such as FOLIO. However, it remains well-suited for rule-based tasks like ProofWriter and ProntoQA. 


\subsection{Free-form Logical Reasoning with LLMs}
The free-form approaches to reasoning rely on LLMs' internal capabilities via various mechanisms to help improve LLM's performance in logical reasoning. For example, prompts that encourage LLMs to solve tasks in a Chain-of-Thought approach is a general technique that enhances LLM's performance \cite{CoF,ZeroShot}. Despite the promising outcomes, this approach falls short when dealing with complex logical reasoning tasks. This limitation stems from the lack of explicit logical grounding and the inherent ambiguous and nuanced nature of natural language. Recent studies have revisited Formal Logic to address this challenge. \citet{FOLIO} shows that incorporating first-order logic (FOL) translations into the context can notably enhance LLM's performance. \citet{logical_solver} emulates the reasoning processes of an automated theorem solver (Pyke) through solving Logical tasks using the tool-based approach and training LLMs on Pyke's reasoning steps. The free-form approach capitalises on the inherent capabilities of LLM to learn complex logical rules. However, this approach solely relies upon LLM's logical reasoning prowess and is susceptible to issues such as hallucinations and taking shortcuts \cite{Shortcut, Hallucinate-Survey}. To address this issue, recent approaches aim to augment LLMs with external symbolic solvers~\cite{satlm, ProglM}.

\subsection{Tool-based Logical Reasoning with LLMs} \label{tool-based approach}
\citet{satlm} and \citet{ProglM} integrated Z3 and Python interpreters with LLMs to tackle various reasoning datasets. \citet{logicLM} expanded upon this by incorporating a broader range of symbolic solvers and employing error-solving self-refinement techniques. However, the rationale behind the adoption of symbolic solvers primarily relied on theoretical definitions rather than empirical performance evaluations. Consequently, there exists a gap in the literature regarding the exploration of the interplay between LLMs, symbolic solvers, and their respective performance characteristics.

The primary advantages of the tool-based approach are: (1) The tasks are now processed with clear logical grounding and unambiguous language. This approach guarantees that the answer is not a product of hallucination or shortcuts, because the symbolic tools will exhaustively process all logical rules in the premise and only execute clear and correct commands. (2) As LLM's translation capability continues to improve, the tool-based approach will be able to solve more complex logical problems, provided they fall within the logical reasoning capacity of symbolic solvers. (3) The tool errors are clearly labeled and displayed (i.e., run-time error messages). This allows the introduction of various error-solving mechanisms like self-refinement \cite{logicLM}. In contrast, it is difficult for the free-form approach to improve upon its current results in the absence of any reliable feedback, specially in the light of recent debates on LLMs self-correction capability~\cite{huang2024large, DBLP:journals/corr/abs-2402-12563}. In this study, errors are isolated into solver-specific errors (e.g., LLM's translation misses a bracket, which causes the solver to throw an error) and parse errors (i.e., Predicate extraction mistakes or LLMs interpreting the logical statements incorrectly, examples of these are shown in Appendix \ref{incorrect}). 

The main disadvantages of the tool-based approach are: (1) This approach does not apply to tasks that do not have a complete reasoning chain. All symbolic solvers require a full chain of logic to reach the correct conclusion. For instance, consider the followint example: \emph{Premise: People like Mark love bbq. Question: Mark is not Human?} Both humans and LLMs can answer this question correctly, but a tool-based approach will fail. This is due to the break in the chain of logic. The term ``Mark is human'' is missing from the premise. Although this term is obvious for humans and LLMs, symbolic solvers require the exact match in predicates to process the task. A detailed discussion of this issue is included in section \ref{complex logic analysis}. (2) Changes in LLMs can cause solver-specific errors.\footnote{For instance, during the experiment stage, we tried to rerun the SatLM experiment on ProofWriter, but the execution rate dropped from 99\% to 20\%. This is caused by GPT3.5 not being able to add a complete bracket to the method Forall()). It is a surprising mistake that continues to happen.} (3) This approach is unforgiving to simple translation errors. While processing logical tasks, Human and LLMs can often bypass errors to some extent and still reach the correct conclusion. However, a tool-based approach requires the LLM to translate tasks flawlessly, even minor mistakes like misusing suffixes (e.g., ``Jompuses(x)'' instead of ``Jompus(x)'') will cause the symbolic solver to throw an error. One of the main focuses of this study is the analysis of how different symbolic tools handle errors caused by LLMs.

\section{Experiments}
\subsection{Experimental Setup}
In our experiments we assess the performance variations of LLM when paired with various symbolic solvers. We evaluate GPT-4o, GPT-3.5-Turbo, Gemini-Pro-1.0, and Command-r-plus integrated with Z3, Pyke, Prover9 on three common logical reasoning benchmarks (introduced shortly). Unlike \citet{logicLM} and other studies, we exclude self-refinement methods and random guessing procedures. In cases where LLM's translation is infeasible, it will not yield an answer, and any specific errors encountered are documented. The only exception is the missing bracket issue for the translation of Z3, as this was not an issue in experiments done in \citet{satlm} and \citet{logicLM}. We use a one-shot demonstration for all experiments. If different solvers are employed to tackle the same dataset, the given prompt problem remains consistent, with the sole variance lying in the solver-specific translations of the prompts. Examples of the prompt are shown in Appendix \ref{Prompts}. We also expand the one-shot experiment for FOLIO to two-shot and four-shot to highlight the impact of additional shots. The primary metrics for evaluation consist of two key factors: the percentage of executable logical formulations (ExecR.), and the overall accuracy (Acc). 
\begin{table*}[t]
\centering
    \resizebox{\textwidth}{!}{%
\begin{tabular}{l|l|cc|cc|cc}
\toprule
  &     &                                                        \multicolumn{2}{c}{Z3}&\multicolumn{2}{c}{Prover9}&\multicolumn{2}{c}{Pyke}\\ \cmidrule(lr){3-4}\cmidrule(lr){5-6}\cmidrule(lr){7-8}
          \textbf{Dataset}          &     \textbf{LLMs}        & ExecR.                                              & Acc.                                                     & ExecR.                                              & Acc.                                                     & ExecR.                                               & Acc.                                                     \\ \cmidrule(lr){1-1}\cmidrule(lr){2-2}\cmidrule(lr){3-3}\cmidrule(lr){4-4}\cmidrule(lr){5-5}\cmidrule(lr){6-6}\cmidrule(lr){7-7}\cmidrule(lr){8-8}
\multirow{3}{*}{ProofWriter} & gpt-4o         & { 75.00\%}                          & {74.17\%}                          & \cellcolor[HTML]{68CBD0}{97.33\%}  & \cellcolor[HTML]{68CBD0}{\color[HTML]{FE0000} 95.67\%}  & {99.83\%}                         & {79.17\%}                          \\  
                   & gpt-3.5-turbo  & { 84.83\%}                          & {82.88\%}                          & {90.67\%}                          & {\color[HTML]{FE0000} 87.00\%}                          & {62.83\%}                          & {53.33\%}                          \\  
                    (Avg. OWA)& gemini-1.0-pro & { 93.00\%}                          & {\color[HTML]{FE0000} 91.00\%}                          & {86.83\%}                          & {62.50\%}                          & {49.33\%}                          & {36.67\%}                          \\  
                   & command-r-plus & { 88.67\%}                          & {\color[HTML]{FE0000} 87.00\%}                          & {61.33\%}                          & {56.66\%}                          & {61.83\%}                          & {51.50\%}                          \\ \cmidrule(lr){2-8}
\multirow{3}{*}{ProofWriter} & gpt-4o         & { 77.83\%}                          & { 77.83\%}                          & {98.00\%}                          & {\color[HTML]{FE0000} 98.00\%}                          & {99.83\%}                         & {87.00\%}                          \\  
                   & gpt-3.5-turbo  & { 88.33\%}                          & { 88.00\%}                          & {94.00\%}                          & {\color[HTML]{FE0000} 93.83\%}                          & {58.17\%}                          & {51.67\%}                          \\  
                   (Avg. CWA)& gemini-1.0-pro & 96.83\%                                                & {\color[HTML]{FE0000} 96.83\%}                         & {84.83\%}                          & { 58.50\%}                          & {42.83\%}                          & {34.17\%}                          \\  
                   & command-r-plus & { 92.50\%}                          & {\color[HTML]{FE0000} 92.50\%}                          & {58.67\%}                          & {58.33\%}                          & {45.33\%}                          & {41.33\%}                          \\ \cmidrule(lr){2-8}
\multirow{4}{*}{PrOntoQA}           & gpt-4o         & { 96.00\%}                          & { 96.00\%}                          & \cellcolor[HTML]{68CBD0}{100.00\%} & \cellcolor[HTML]{68CBD0}{\color[HTML]{FE0000} 100.00\%} & \cellcolor[HTML]{68CBD0}{100.00\%} & \cellcolor[HTML]{68CBD0}{\color[HTML]{FE0000} 100.00\%} \\  
                   & gpt-3.5-turbo  & { 95.50\%}                          & {\color[HTML]{FE0000} 93.49\%}                          & {85.50\%}                          & {63.50\%}                          & {99.50\%}                          & {72.50\%}                          \\  
                   & gemini-1.0-pro & \cellcolor[HTML]{68CBD0}{ 100.00\%} & \cellcolor[HTML]{68CBD0}{\color[HTML]{FE0000} 100.00\%} & {100.00\%}                         & {97.50\%}                          & \cellcolor[HTML]{68CBD0}{100.00\%} & \cellcolor[HTML]{68CBD0}{\color[HTML]{FE0000} 100.00\%} \\  
                   & command-r-plus & { 93.00\%}                          & { 87.00\%}                          & {64.50\%}                          & {46.50\%}                          & {96.50\%}                          & {\color[HTML]{FE0000} 92.00\%}                          \\\cmidrule(lr){2-8}
\multirow{4}{*}{FOLIO}     & gpt-4o         & { 40.00\%}                          & { 36.00\%}                          & 84.00\%                                                 & {\color[HTML]{FE0000} 66.50\%}                          & {\ding{55}}                                 & {\ding{55}}                                 \\  
                  & gpt-3.5-turbo  & { 29.00\%}                          & { 24.49\%}                          & {61.00\%}                          & {\color[HTML]{FE0000} 39.99\%}                          & {\ding{55}}                                 & {\ding{55}}                                 \\  
                   & gemini-1.0-pro & { 31.00\%}                          & { 25.50\%}                          & {67.50\%}                          & {\color[HTML]{FE0000} 50.00\%}                          & {\ding{55}}                                 & {\ding{55}}                                 \\  
                   & command-r-plus & { 25.50\%}                          & {19.00\%}                          & {50.50\%}                          & {\color[HTML]{FE0000} 32.50\%}                          & {\ding{55}}                                 & {\ding{55}}                                 \\        \specialrule{2.5pt}{1pt}{1pt}
\multirow{4}{*}{Combined}            & gpt-4o         & { 74.31\%}                          & 73.50\%                                                 & \cellcolor[HTML]{68CBD0}94.06\%                         & \cellcolor[HTML]{68CBD0}{\color[HTML]{FF0000} 91.71\%}  & 99.86\%                                                 & 85.50\%                                                 \\  
                   & gpt-3.5-turbo  & { 80.50\%}                          & 78.83\%                                                 & 87.56\%                                                 & {\color[HTML]{FF0000} 80.75\%}                          & 66.07\%                                                 & 55.36\%                                                 \\  
                   & gemini-1.0-pro & { 87.56\%}                          & {\color[HTML]{FF0000} 86.12\%}                          & 85.31\%                                                 & 63.81\%                                                 & 53.79\%                                                 & 44.64\%                                                 \\  
                   & command-r-plus & { 82.75\%}                          & {\color[HTML]{FF0000} 80.56\%}                          & 59.38\%                                                 & 53.00\%                                                 & 60.64\%                                                 & 52.93\%                                                 \\ \bottomrule
\end{tabular}}
\caption{Accuracy and execution rate of 1-shot experiments done with gpt-4o, gpt-3.5-turbo, gemini-pro-1.0 and command-r-plus on 3 Datasets. Results for Proofwriter Open and Closed World Assumptions (OWA and CWA) are averaged over depths (Depth 2, 3, and 5). We present the percentage of executable logical formulations (ExecR.) together with the overall accuracy (Acc.). \ding{55}: the tool was unable to solve this dataset. The numbers highlighted in red color represent the highest accuracy between the 3 chosen tools.}
\label{Main Table}
\end{table*}
\paragraph{Data} The 3 benchmarks are introduced shortly and examples are included in Appendix \ref{Dataset Examples}. We limit the test set size to 200 for cost reason. \textbf{PrOntoQA} \cite{ProntoQA} is a synthetic dataset created to analyze the capacity of LLMs for deductive reasoning. We use the hardest fictional characters version and the hardest 5-hop subset for evaluation. PrOntoQA only has questions in the close world setting (i.e., True/False only). We include this dataset in the experiment to compare natural and fictional settings, as it has a similar level of logical difficulty to ProofWriter. \textbf{ProofWriter} \cite{ProofWriter} is a commonly used dataset for deductive logical reasoning. Compared with PrOntoQA, the problems are expressed in a more naturalistic language form.  We evaluate 6 different variations of ProofWriter. We use both open-world (OWA) and close-world assumptions (CWA), including depth-2, depth-3, and depth-5 (i.e., each part requiring 2, 3, and 5 hops of reasoning). To ensure a fair evaluation, we control all datasets to have a uniform distribution of True, False, and Unknown (if applicable) answers. \textbf{FOLIO} \cite{FOLIO} is a difficult expert-written dataset for first-order logical reasoning. The problems are mostly aligned with real-world knowledge and expressed in natural flowing language. Tackling its questions demands adeptness in complex first-order logic reasoning. Pyke is unable to solve FOLIO, this is due to the lack of a built-in function for the exclusive disjunction (i.e., either-or). In contrast, Prover9 and Z3 offer a built-in function to handle this logic seamlessly. 

\subsection{Main Results}\label{sec:mainres}
We report the results of the tool-based reasoning approach experiments in Table \ref{Main Table}. 
%
%
Different LLMs exhibit varying preferences for tools. For datasets with simpler logical complexity, GPT models tend to favor Prover9, while Gemini and Command R+ models perform significantly better using Z3. Pyke is only competitive in solving PrOntoQA and is unable to solve datasets like FOLIO and performs significantly worse on ProofWriter. Pyke's primary issue is the low and inconsistent executable rate. 
\begin{figure*}[t]
    \centering
    \includegraphics[width=1\linewidth]{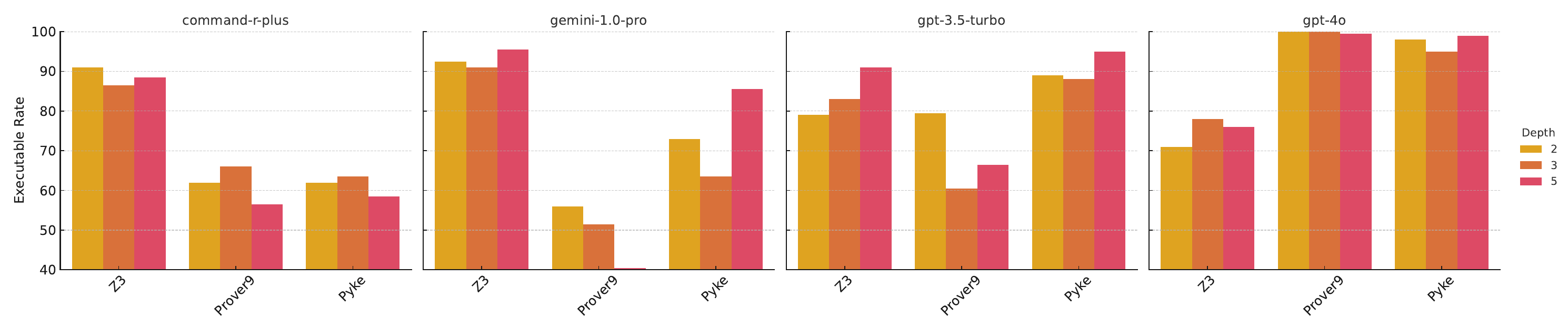}
    \caption{Executable Rate for different LLM-Tool combinations, for depth 2, 3, 5 of the ProofWriter Open World Assumption (OWA). Similar trend exists for the Close World Assumption (CWA).}
    \label{fig: Proof OWA ACC}
\end{figure*}
According to Table \ref{Average Main table}, without considering the option of LLMs, Prover9 performs better for the FOLIO dataset, Z3 performs better on other datasets. Both Z3 and Prover9 have their distinct advantages. Prover9's programming language, which closely resembles the language of First-Order Logic (FOL), contributes to its higher execution rate. 
{The Pearson correlation coefficient  between executable rate and accuracy across all LLMs (Prover9, Z3, and Pyke have, 0.98, 0.82, 0.94, respectively\footnote{\text{p-values:} $1.02 \times 10^{-18},\  5.37 \times 10^{-7},\  6.1 \times 10^{-12}$}) indicate an almost linear dependence between execution success and accuracy for Prover9. The lower correlation for Z3 highlights the discrepancy between writing executable codes for Z3, and the accuracy of those codes.}

\paragraph{Natural vs. Fictional}\label{N vs F}
We compare the performance of ProntoQA Depth 5 and ProofWriter CWA Depth 5 to investigate how different symbolic solvers affect the performance of tool-augmented LLMs in natural versus fictional world settings. The main difference between the datasets is that PrOntoQA uses fictional characters (i.e., imaginary characters like Jompus and Wompus), while ProofWriter is expressed in more naturalistic language. \citet{ProntoQA_OOD} have shown that real-world knowledge helps LLMs in reasoning more effectively, a fictional world setting decreases LLM's logical performance. On average, Prover9's performance is most aligned with this observation. The executable rate on average decreases for all LLMs, and average accuracy drops by 1.38\% in a fictional setting. Both Z3 and Pyke's overall accuracy increased by 6.62\% and 30.87\%. This shows that while using Z3 and Prover9, fictional wording helps LLMs in generating consistent and correct translations. Overall, in a fictional setting, Pyke's performance is significantly boosted. Meanwhile, GPT-3.5-Turbo shifts its preference from Prover9 to Z3, and Command R+ changes its preference to Pyke. We speculate the nuance in results to be reflective of potential interference between commonsense knowledge and fictional statements.

\paragraph{Depth} \label{depth analysis}
The relaiton between depth and executable rate is somewhat mixed, specially between depth 2 and 3. While for command-r-plus we observe a general decay in performance (i.e., between depth 2 and 5) across all tools, both GPT models and Gemini exhibit resilence to depth, with performance even improving across most tools (except for Prover9). This observation highlights the robustness of translation-based approaches (i.e., using LLMs for translation and tools for solving) in handling various complexities, while prior findings reported the reasoning ability of LLMs (alone) generally diminish as the number of reasoning hops increases~\cite{FOLIO}. 

\paragraph{Demonstration Shots} \label{complex logic analysis}
We present the statistics of the FOLIO dataset in varying number of shots in Table \ref{Main FOLIO table}. Prover9 achieves the best performance, while Z3 struggles with execution rate. The best result for FOLIO was 66.5\%, which is achieved with 1 shot prompting using GPT-4o and Prover9. The primary factors that limit the execution rate performance on FOLIO are: (1) some natural wordings in FOLIO make it difficult for predicate extraction. For example, GPT4o interpreted the term "Eastern wild turkey" as two separate terms "Eastern(x)" and "WildTurkey(x)", but "Eastern(x)" has no meaning and the predicate should be extracted as EasternWildTurkey(x). (2) FOLIO is annotated by humans and thus assumes a degree of commonsense, this presents incomplete reasoning chains and ambiguous sentences. As shown in \ref{FOLIO Incorrect Interpretation}, GPT-3.5-Turbo incorrectly translated the statement ``Marvin cannot be from Earth and from Mars.'' into ``Not(And(FromEarth(marvin), FromMars(marvin)))'', which entails Marvin is not from Earth and not from Mars. The simple fix is just to change Not() into Xor(). This problem was caused by the inherently ambiguous nature of the natural language. (3) there is a limitation to learning by increasing the number of shots. Specifically, GPT-4o and Prover9's parse errors increased with a higher number of shots, as shown in Table \ref{Main FOLIO table}. Overall, while Prover9 can solve a greater number of questions, Z3 shows significant potential in addressing FOLIO. This is due to Z3's error-display capabilities, which are essential for continuous improvement.
\begin{table}[t]
\resizebox{\columnwidth}{!}{%
\begin{tabular}{l|cc|cc}
\toprule
& \multicolumn{2}{c}{Z3} & \multicolumn{2}{c}{Prover9}\\ \cmidrule(lr){2-3}\cmidrule(lr){4-5}
& \multicolumn{1}{c}{ExecR.} & \multicolumn{1}{c}{Acc.}    & \multicolumn{1}{c}{ExecR.} & \multicolumn{1}{c}{Acc.}         \\ \midrule
                 \multicolumn{5}{c}{\textbf{GPT-4o}}\\ \midrule
$k=1$         & 40\%                           & { 36\%}      & { 84\%} & 66.5\%                               \\
$k=2$         & 50.5\%                         & { 40.96\%}   & { 74.5\%}    & 58\%                               \\
$k=4$         & 51\%                           & 39.5\%                          & { 77\%}    & { 62\%}        \\  \midrule
                 \multicolumn{5}{c}{\cellcolor{gray!30}\textbf{GPT-3.5-Turbo}}\\  \midrule
$k=1$  & 29\%                           & { 24.49\%}   & { 61\%}    & 39.99\%                               \\
$k=2$   & 37\%                           & { 31\%}   & { 58\%}    & 40.5\%                               \\
$k=4$         & 48\%                           & { 36.5\%}   & { 65\%}    & 44.5\%                               \\  \midrule
                 \multicolumn{5}{c}{\cellcolor{gray!30}\textbf{Gemini-1.0-Pro}}\\  \midrule
$k=1$ & 31\%                           & { 25.5\%}   & { 67.5\%}  & 50.00\%                               \\
$k=2$   & 47.5\%                         & { 39\%}    & { 60.5\%}  & 38\% \\
$k=4$         & 48\%                           & { 36.5\%}   & { 65.5\%}  & 44\%                               \\    \midrule
                 \multicolumn{5}{c}{\cellcolor{gray!30}\textbf{Command-R-Plus}}\\ \midrule
$k=1$ & 25.50\%                        & { 19.00\%}   & { 50.50\%} & 32.50\%                                \\
$k=2$  & 33.5\%                        & { 26.5\%}   & { 42.5\%} & 29.5\%    \\
$k=4$         & 42\%                        & { 32.5\%}   & { 60.5\%} &46\%    \\ \bottomrule
\end{tabular}}
\caption{The effect of varying number of shots ($k=1,2,4$) on accuracy and executable rates under GPT-4o, GPT-3.5-turbo, Gemini-1.0-pro and command-r-plus on FOLIO. We present the percentage of executable logical formulations (ExecR.) together with the overall accuracy (Acc.).}
\label{Main FOLIO table}
\end{table}
\section{Analysis} \label{Error analysis}
As indicated by the executable rate in Table \ref{Main Table}, LLMs generally find it easier to produce executable logical formulations for Prover9. This is attributed to its foundation in FOL-based programming language, which most large language models (LLMs) are familiar with as a form of logical formulation. While GPT models are more successful at converting these logical formulations into accurate results, Gemini-1.0-pro and Command R+ face challenges in achieving similar accuracy. This is an issue because an executable formulation cannot provide feedback when an incorrect result is given. This hinders further improvement and self-refinement. Z3 does not have this issue. Its executable rate is a reflection of its accuracy. Moreover, Z3's programming language closely aligns with Python, offering a unique advantage in error displaying and further improvement. Z3 is also a flexible tool that allows the inclusion of self-defined complex logical rules like "XorAnd()" (i.e., a combination of the rule "Either or" and "And".). This capability is particularly useful for addressing complex reasoning datasets like FOLIO. We did not define such a rule during our experiment but this capability should be considered in further studies. 

Non-executable logical formulations can be categorized into \emph{parse errors} and \emph{execution errors}. Additionally, for Z3, there is a separate category known as \emph{execution exceptions}.

\begin{figure*}[t]
    \centering
    \includegraphics[trim={3.5cm 2.5cm 2.5cm 2.5cm},clip,width=\linewidth]{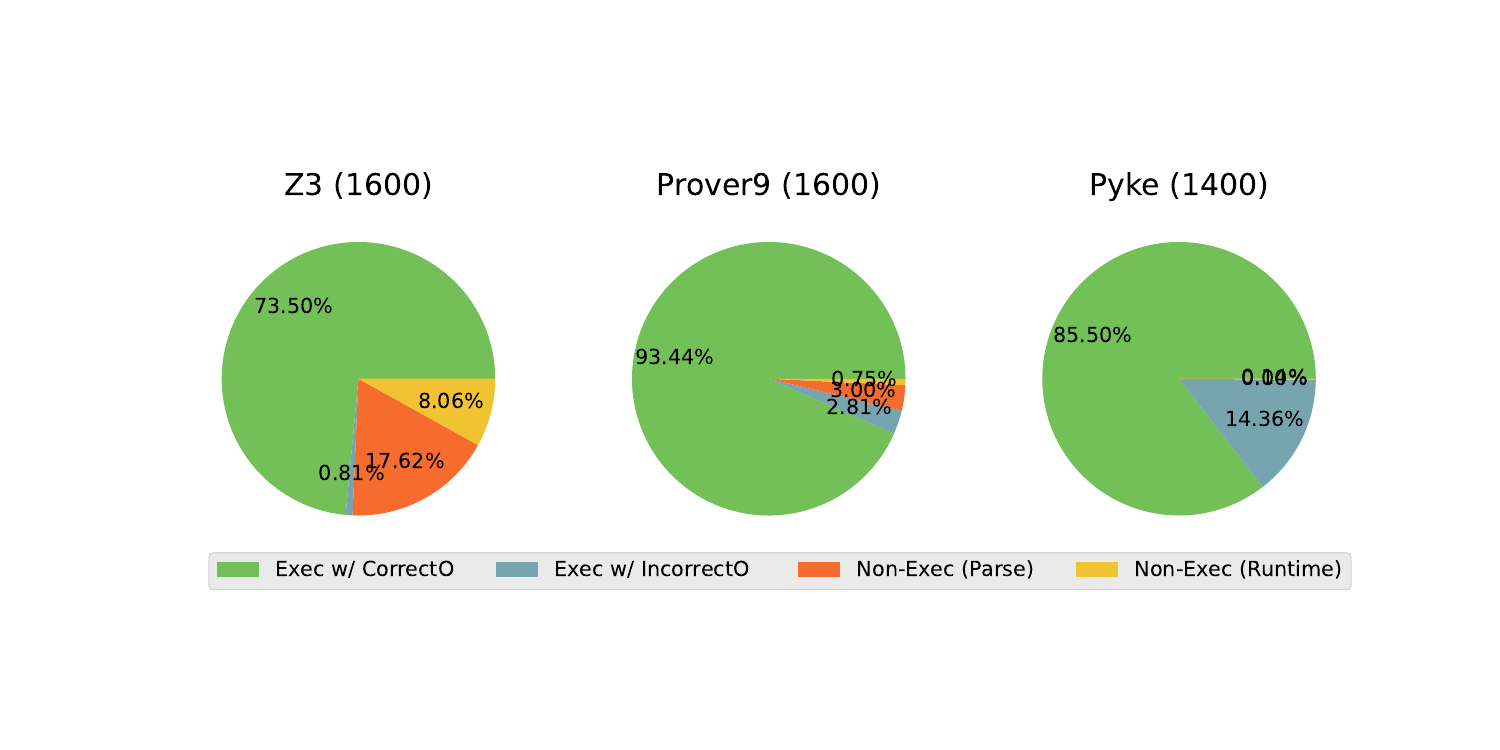}
    \caption{The proportion of various executable and non-executable instances per each tool for GPT4o. Note, Pyke does not include FOLIO (hence 1400 instances compared to Z3 and Prover 9). The \emph{Exec w/ CorrectO}, and \emph{Exec w/ IncorrectO} denote Executable translations that lead to correct, and incorrect outputs once executed by the tool. The \emph{Non-exec (Parse) or (Runtime)} denote the non-executable translations which are either due to parsing error or other potential runtime issues.}
    \label{fig:pies}
\end{figure*}

\begin{itemize}[leftmargin=*]
\setlength\itemsep{0em}
\item \textbf{\emph{parse error}} refers to the mistakes identified by the parser. Through the prompt, we have predefined a set of instructions and logical rules that LLMs can use. However, when LLMs hallucinate and generate logical rules or code that do not exist in the solver, the parser will detect these discrepancies and throw an parse error. This error indicates the LLM's inability to adhere to the one-shot prompt, resulting in methods or code that the parser cannot process. For instance, using Exist() instead of Exists() for Z3 is an example of such an error.
\item \textbf{\emph{execution error}} occurs when the solver encounters given facts that are inconsistent, predicates that are defined wrong, or when there are solver-specific syntax errors. This type of error can be resolved through self-refinement, as the errors are explicitly displayed. We call this run-time error.
\item \textbf{\emph{execution exception}} is a special case for Z3, where the solver runs both the original conclusion and the negation of the same conclusion but receives true as the answer in both cases. This indicates that the facts are inconsistent. We combined these errors into run-time errors for Figure~\ref{fig:pies} Z3 visualisation.
\end{itemize}
As shown in Figure \ref{fig:pies}, for GPT4o, while Pyke produced 3 execution errors on easier logical reasoning datasets in total, its high execution rate did not translate to high accuracy. Predominately Prover9 and Z3's error is a parse error, with execution error controlled at around 8 questions. In addition, all non-executable questions are different, there are no common questions that all 3 solvers find difficult to solve. For FOLIO, the execution error increases, and the parse error drops significantly. Challenging datasets, such as FOLIO, encompass a larger number of unseen, complex logical rules and more intricate predicates, which result in higher error rates during translation by LLMs. Additionally, there is an increasing number of questions that both solvers are unable to process. This suggests that both solvers find around 25-30\% of questions hard to solve.

\section{Conclusion}
In this study, we investigated and compared the performance of LLMs combined with three widely used symbolic solvers to closely examine how each solver influences the performance of tool-augmented LLMs in logical reasoning. Our experiments demonstrated that the choice of tools (i.e., Z3, Pyke, Prover9) has a significant impact on the downstream performance across various benchmarks and LLMs.

\section*{Limitations}
The tool-based approach to logical reasoning is limited to deductive reasoning datasets with a complete reasoning chain. This constraint arises from the inherent nature of symbolic solvers. A potential solution is for LLMs to generate the missing segments of the reasoning chain. Additionally, black-box LLMs can exhibit inconsistencies, producing results that change in the course of time. For instance, during our experiment, GPT-3.5-Turbo consistently failed to add a closing bracket to the method "Forall()", while Command R+ failed to include an opening bracket. This was not an issue for \citet{logicLM} and \citet{satlm} (or at least was not reported in their papers). We limited our use of solvers to their built-in functions. To enhance the performance of each tool, more unique logical combinations can be integrated and implemented. For example, Z3 is a flexible tool that allows the inclusion of rules such as "Male(x) == Not(Female(x))".  There is further potential to include more defined complex logical rules that can make LLM translation easier. 
\bibliography{main}
\newpage
\appendix
\onecolumn
\section{Appendix}
\label{sec:appendix}


\subsection{Dataset Examples} \label{Dataset Examples}
\begin{boxK}
\textbf{ProofWriter}\\
Example: ProofWriter Depth 5 Open World Assumption Q774\\

Problem:\\
The bald eagle is blue. The bald eagle is kind. The bald eagle likes the cat. The bald eagle does not visit the tiger. The cat chases the mouse. The cat is green. The cat likes the bald eagle. The cat likes the mouse. The cat does not like the tiger. The mouse likes the cat. The tiger chases the cat. The tiger chases the mouse. The tiger is red. The tiger likes the cat. The tiger visits the cat. The tiger visits the mouse. If something likes the bald eagle then it is blue. If something visits the bald eagle and it visits the cat then the bald eagle is red. If something chases the mouse then it visits the cat. If something is blue then it chases the tiger. If something visits the cat and the cat chases the tiger then the tiger likes the bald eagle. If something likes the tiger then the tiger likes the bald eagle. If something chases the mouse then it visits the mouse.\\

Question:\\
Based on the above information, is the following statement true, false, or unknown? \\
The cat does not like the mouse.\\

Answer: False\\
\end{boxK}
\begin{boxK}
\textbf{PrOntoQA}\\
Example: ProntoQA Q3\\

Problem:\\
Vumpuses are floral. Vumpuses are tumpuses. Tumpuses are brown. Each tumpus is a wumpus. Wumpuses are small. Each wumpus is a rompus. Each zumpus is metallic. Every rompus is happy. Rompuses are impuses. Each impus is amenable. Each impus is a dumpus. Every dumpus is not metallic. Dumpuses are numpuses. Each numpus is bitter. Each numpus is a jompus. Every jompus is cold. Each jompus is a yumpus. Wren is a tumpus.
Question:\\
Is the following statement true or false? \\
Wren is not metallic.\\

Answer: True\\
\end{boxK}

\begin{boxK}
\textbf{FOLIO}\\
Example: FOLIO dev Q1\\

Problem:\\
If people perform in school talent shows often, then they attend and are very engaged with school events. People either perform in school talent shows often or are inactive and disinterested members of their community. If people chaperone high school dances, then they are not students who attend the school. All people who are inactive and disinterested members of their community chaperone high school dances. All young children and teenagers who wish to further their academic careers and educational opportunities are students who attend the school. Bonnie either both attends and is very engaged with school events and is a student who attends the school, or she neither attends and is very engaged with school events nor is a student who attends the school.\\

Question:\\
Based on the above information, is the following statement true, false, or uncertain? \\
If Bonnie is either both a young child or teenager who wishes to further her academic career and educational opportunities and chaperones high school dances or neither is a young child nor teenager who wishes to further her academic career and educational opportunities, then Bonnie is either a student who attends the school or is an inactive and disinterested member of the community.\\

Answer: True\\
\end{boxK}
\subsection{Prompts} \label{Prompts}
\paragraph{ProofWriter Prompts for Z3 Solver} One-shot demonstration 
\begin{boxK}
Given a problem description and a question. The task is to parse the problem and the question into Python Z3 solver.

------

Problem:

Anne is quiet. 
Erin is furry. 
Erin is green. 
Fiona is furry. 
Fiona is quiet. 
Fiona is red. 
Fiona is rough. 
Fiona is white. 
Harry is furry. 
Harry is quiet. 
Harry is white. 
Young people are furry. 
If Anne is quiet then Anne is red. 
Young, green people are rough. 
If someone is green then they are white. 
If someone is furry and quiet then they are white. 
If someone is young and white then they are rough. 
All red people are young.

Question:

Based on the above information, is the following statement true, false, or unknown? Anne is white.
\#\#\#

\# solution in Python:

def solution():

\# Anne is quiet.

quiet(Anne)

\# Erin is furry.

furry(Erin)

\# Erin is green.

green(Erin)

\# Fiona is furry.

furry(Fiona)

\# Fiona is quiet.

quiet(Fiona)

\# Fiona is red.

red(Fiona)

\# Fiona is rough.

rough(Fiona)

\# Fiona is white.

white(Fiona)

\# Harry is furry.

furry(Harry)

\# Harry is quiet.

quiet(Harry)

\# Harry is white.

white(Harry)

\# Young people are furry.

ForAll([x], Implies(young(x), furry(x)))

\# If Anne is quiet then Anne is red.

Implies(quiet(Anne), red(Anne))

\# Young, green people are rough.

ForAll([x], Implies(And(young(x), green(x)), rough(x)))

\# If someone is green then they are white.

ForAll([x], Implies(green(x), white(x)))

\# If someone is furry and quiet then they are white.

ForAll([x], Implies(And(furry(x), quiet(x)), white(x)))

\# If someone is young and white then they are rough.

ForAll([x], Implies(And(young(x), white(x)), rough(x)))

\# All red people are young.

ForAll([x], Implies(red(x), young(x)))

\# Question: the following statement true, false, or unknown? Anne is white.

return white(Anne)
\end{boxK}
\paragraph{ProofWriter Prompts for Prover9} One shot demonstration for LLM

\begin{boxK}
Given a problem description and a question, the task is to parse the problem and the question into first-order logic formulas. The grammar of the first-order logic formula is defined as follows:
\begin{enumerate}
    \item Logical conjunction of $\text{expr1}$ and $\text{expr2}$: $\text{expr1} \land \text{expr2}$
    \item Logical disjunction of $\text{expr1}$ and $\text{expr2}$: $\text{expr1} \lor \text{expr2}$
    \item Logical exclusive disjunction of $\text{expr1}$ and $\text{expr2}$: $\text{expr1} \oplus \text{expr2}$
    \item Logical negation of $\text{expr1}$: $\neg \text{expr1}$
    \item $\text{expr1}$ implies $\text{expr2}$: $\text{expr1} \rightarrow \text{expr2}$
    \item $\text{expr1}$ if and only if $\text{expr2}$: $\text{expr1} \leftrightarrow \text{expr2}$
    \item Logical universal quantification: $\forall x$
    \item Logical existential quantification: $\exists x$
\end{enumerate}
\text{Problem} \\
Anne is quiet. 
Erin is furry. 
Erin is green. 
Fiona is furry. 
Fiona is quiet. 
Fiona is red. 
Fiona is rough. 
Fiona is white. 
Harry is furry. 
Harry is quiet. 
Harry is white. 
Young people are furry. 
If Anne is quiet then Anne is red. 
Young, green people are rough. 
If someone is green then they are white. 
If someone is furry and quiet then they are white. 
If someone is young and white then they are rough. 
All red people are young.

Question:

Based on the above information, is the following statement true, false, or unknown? Anne is white.

\#\#\# \\
\text{Predicates} \\
\text{quiet(x)} :::  \text{x is quiet.}\\
\text{furry(x)} ::: \text{x is furry.} \\
\text{green(x)} ::: \text{x is green.}\\
\text{red(x)} ::: \text{x is red.}\\
\text{rough(x)} ::: \text{x is rough.}\\
\text{white(x)} ::: \text{x is white.}\\
\text{young(x)} ::: \text{x is young}\\
\text{Premises} \\
\text{quiet(Anne)} :::  \text{Anne is quiet.}\\
\text{furry(Erin)} ::: \text{Erin is furry.} \\
\text{green(Erin)} ::: \text{Erin is green.}\\
\text{furry(Fiona)} ::: \text{Fiona is furry.} \\
\text{quiet(Fiona)} ::: \text{Fiona is quiet.} \\
\text{red(Fiona)} ::: \text{Fiona is red.}\\
\text{rough(Fiona)} ::: \text{Fiona is rough.}\\
\text{white(Fiona)} ::: \text{Fiona is white.}\\
\text{furry(Harry)} ::: \text{Harry is furry.}\\
\text{quiet(Harry)} ::: \text{Harry is quiet.}\\
\text{white(Harry)} ::: \text{Harry is white.}\\
$\forall x (\text{young}(x) \rightarrow \text{furry}(x)) \,:::\, \text{Young people are furry.}$ \\
$(\text{quiet(Anne)} \rightarrow \text{red(Anne)}) ::: \text{If Anne is quiet then Anne is red.}$\\
$\forall x (\text{young}(x) \land \text{green}(x) \rightarrow \text{rough}(x))$ ::: \text{Young, green people are rough.} \\
$\forall x (\text{green}(x) \rightarrow \text{white}(x)) ::: \text{If someone is green then they are white.} $\\
$\forall x ((\text{furry}(x) \land \text{quiet}(x)) \rightarrow \text{white}(x)) ::: \text{If someone is furry and quiet then they are white.}$ \\
$\forall x ((\text{young}(x) \land \text{white}(x)) \rightarrow \text{rough}(x)) ::: \text{If someone is young and white then they are rough.}$\\
$\forall x (\text{red}(x) \rightarrow \text{young}(x)) ::: \text{All red people are young.}$

\text{Conclusion:}\\
\text{white(Anne)}

\end{boxK}

\paragraph{ProofWriter Prompts for Pyke Solver} One-shot demonstration

\begin{boxK}
Task Description: You are given a problem description and a question. The task is to: \\
1) define all the predicates in the problem \\
2) parse the problem into logic rules based on the defined predicates \\
3) write all the facts mentioned in the problem \\
4) parse the question into the logic form \\
\text{Problem:} \\
Anne is quiet. 
Erin is furry. 
Erin is green. 
Fiona is furry. 
Fiona is quiet. 
Fiona is red. 
Fiona is rough. 
Fiona is white. 
Harry is furry. 
Harry is quiet. 
Harry is white. 
Young people are furry. 
If Anne is quiet then Anne is red. 
Young, green people are rough. 
If someone is green then they are white. 
If someone is furry and quiet then they are white. 
If someone is young and white then they are rough. 
All red people are young.

Question:

Based on the above information, is the following statement true, false, or unknown? Anne is white.\\
\#\#\# \\
\text{Predicates:} \\
\text{quiet(\$x, bool)} :::  \text{x is quiet.}\\
\text{furry(\$x, bool)} ::: \text{x is furry.} \\
\text{green(\$x, bool)} ::: \text{x is green.}\\
\text{red(\$x, bool)} ::: \text{x is red.}\\
\text{rough(\$x, bool)} ::: \text{x is rough.}\\
\text{white(\$x, bool)} ::: \text{x is white.}\\
\text{young(\$x, bool)} ::: \text{x is young}\\
\text{Facts:} \\
\text{quiet(Anne, True)} :::  \text{Anne is quiet.}\\
\text{furry(Erin, True)} ::: \text{Erin is furry.} \\
\text{green(Erin, True)} ::: \text{Erin is green.}\\
\text{furry(Fiona, True)} ::: \text{Fiona is furry.} \\
\text{quiet(Fiona, True)} ::: \text{Fiona is quiet.} \\
\text{red(Fiona, True)} ::: \text{Fiona is red.}\\
\text{rough(Fiona, True)} ::: \text{Fiona is rough.}\\
\text{white(Fiona, True)} ::: \text{Fiona is white.}\\
\text{furry(Harry, True)} ::: \text{Harry is furry.}\\
\text{quiet(Harry, True)} ::: \text{Harry is quiet.}\\
\text{white(Harry, True)} ::: \text{Harry is white.}\\
\text{young}(\$x, True) $>>>$ \text{furry}(\$x, True)) \,:::\, \text{Young people are furry.} \\
\text{quiet(Anne, True)} $>>>$ \text{red(Anne, True)}) ::: \text{If Anne is quiet then Anne is red.}\\
\text{young}(\$x, True) \&\& \text{green}(\$x, True) $>>>$ \text{rough}(\$x, True) ::: \text{Young, green people are rough.} \\
\text{green}(\$x, True) $>>>$ \text{white}(\$x, True) ::: \text{If someone is green then they are white.} \\
\text{furry}(\$x, True) \&\& \text{quiet}(\$x, True) $>>>$ \text{white}(\$x, True) \\::: \text{If someone is furry and quiet then they are white.} \\
\text{young}(\$x, True) \&\& \text{white}(\$x, True) $>>>$ \text{rough}(\$x, True)\\ ::: \text{If someone is young and white then they are rough.}\\
\text{red}(\$x, True)  $>>>$ \text{young}(\$x, True) ::: \text{All red people are young.} \\
\text{Query:}\\
\text{white(Anne)}
\end{boxK}

\subsection{Incorrect Example Generation}\label{incorrect}
The following section includes classic Incorrect translations, more incorrect translations can be found in Processed\_Datasets in \url{https://github.com/Mattylam/Logic\_Symbolic\_Solvers\_Experiment}
\begin{boxK}
\textbf{Example 1: Prover9 PrOntoQA\_2 Parse error}

..........

$\forall$x(Dumpus(x) $\rightarrow$ Happy(x)) ::: Every dumpus is happy.\\
$\forall$x(Dumpus(x) $\rightarrow$ Vumpus(x)) ::: Each dumpus is a vumpus.\\
$\forall$x(Vumpus(x) $\rightarrow$ Bright(x)) ::: Vumpuses are bright.\\
$\forall$x(Vumpus(x) $\rightarrow$ Jompus(x)) ::: Every vumpus is a jompus.\\
$\forall$x(Jompus(x) $\rightarrow$ Large(x)) ::: Jompuses are large.\\
\textcolor{red}{Stella is a yumpus.}\\
Conclusion:\\
Bright(Stella) ::: Stella is bright.\\

\textcolor{brown}{Analysis: LLM failed to perform all translations, "Stella is yumpus" was not translated, thus system throws a parse error}
\end{boxK}
\begin{boxK} \label{FOLIO Incorrect Interpretation}
\textbf{Example 2: Z3 FOLIO\_dev\_27 Incorrect Interpretation}
\\

Problem: All aliens are extraterrestrial. If someone is from Mars, then they are aliens. No extraterrestrial is human. Everyone from Earth is a human. \textcolor{red}{Marvin cannot be from Earth and from Mars.} If Marvin is not from Earth, then Marvin is an extraterrestrial.\\
Question: Based on the above information, is the following statement true, false, or uncertain? Marvin is an alien.\\
Answer: False\\
Predicted Answer: Unknown\\
Translation from LLM:\\
........\\
\# All aliens are extraterrestrial.\\
ForAll([x], Implies(Alien(x), Extraterrestrial(x)))\\
\# If someone is from Mars, then they are aliens.\\
ForAll([x], Implies(FromMars(x), Alien(x)))\\
\# No extraterrestrial is human.\\
ForAll([x], Implies(Extraterrestrial(x), Not(Human(x))))\\
\# Everyone from Earth is a human.\\
ForAll([x], Implies(FromEarth(x), Human(x)))\\
 \textcolor{red}{\# Marvin cannot be from Earth and from Mars.}\\
 \textcolor{red}{Not(And(FromEarth(marvin), FromMars(marvin)))}\\
\# If Marvin is not from Earth, then Marvin is an extraterrestrial.\\
Implies(Not(FromEarth(marvin)), Extraterrestrial(marvin))\\
\# Question: Marvin is an alien.\\
return Alien(marvin)\\

\textcolor{brown}{Analysis: LLM incorrectly interpreted the logic of the statement. The correct translation should be Xor(And(FromEarth(marvin), FromMars(marvin))), the wording of the statement makes it ambiguous for LLMs to interpret. If the statement was, Marvin is either from Earth or from Mars, LLMs would have been able to translate it.}

\end{boxK}

\begin{boxK} \label{FOLIO Difficult predicate extraction}
\textbf{Example 3: Z3 FOLIO\_dev\_16 Difficult predicate extraction}

Problem: All rabbits are cute. \textcolor{red}{Some turtles exist.} An animal is either a rabbit or a squirrel. If something is skittish, then it is not still. All squirrels are skittish. Rock is still. \\
Question: Based on the above information, is the following statement true, false, or uncertain? Rock is a turtle.
Answer: Unknown\\
Error: ExecutionError NameError: name 'Exist' is not defined. Did you mean: 'exit'?"\\
Translation from LLM:\\
........\\
All rabbits are cute.\\
ForAll([x], Implies(Rabbit(x), Cute(x)))\\
\# Some turtles exist.\\
\textcolor{red}{Exist([x], Turtle(x))}\\
\# An animal is either a rabbit or a squirrel.\\
ForAll([x], Or(Rabbit(x), Squirrel(x)))\\
\# If something is skittish, then it is not still.\\
ForAll([x], Implies(Skittish(x), Not(Still(x))))\\
\# All squirrels are skittish.\\
ForAll([x], Implies(Squirrel(x), Skittish(x)))\\
\# Rock is still.\\
Still(Rock)\\
\# Question: Rock is a turtle.\\
return Turtle(Rock)\\

\end{boxK}

\subsection{How Z3 tackles BoardgameQA}\label{BoardgameQA Z3}
\begin{boxK}
A few players are playing a boardgame. The current state of the game is as follows.
The cat is named Meadow.
The cat published a high-quality paper.
The gecko winks at the eagle.
The puffin is named Peddi. 
And the rules of the game are as follows. 
Rule1: Regarding the cat, if it has a high-quality paper, then we can conclude that it prepares armor for the canary. 
Rule2: If you are positive that you saw one of the animals rolls the dice for the eagle, you can be certain that it will also proceed to the spot right after the sun bear. 
Rule3: If the cat prepares armor for the canary and the cricket gives a magnifier to the canary, then the canary will not remove one of the pieces of the aardvark. 
Rule4: If the cat has a name whose first letter is the same as the first letter of the puffin's name, then the cat does not prepare armor for the canary. 
Rule5: The canary removes one of the pieces of the aardvark whenever at least one animal proceeds to the spot right after the sun bear. 
Rule6: Regarding the cat, if it has something to sit on, then we can conclude that it does not prepare armor for the canary. 
Rule3 is preferred over Rule5. 
Rule4 is preferred over Rule1. 
Rule6 is preferred over Rule1. \\
Based on the game state and the rules and preferences, does the canary remove from the board one of the pieces of the aardvark?\\

\# solution in Python:\\
def solution():\\
\# rules\\
\# Rule1: Regarding the cat, if it has a high-quality paper, then we can conclude that it prepares armor for the canary.\\
Rule1 = Implies(has\_high\_quality\_paper(cat), prepare\_armor(cat, canary))\\    
\# Rule2: If you are positive that you saw one of the animals rolls the dice for the eagle, you can be certain that it will also proceed to the spot right after the sun bear.\\
Rule2 = ForAll([x], Implies(roll\_dice(x, eagle), proceed\_to\_spot\_right\_after(x, sun\_bear)))\\
\# Rule3: If the cat prepares armor for the canary and the cricket gives a magnifier to the canary, then the canary will not remove one of the pieces of the aardvark.\\
Rule3 = Implies(And(prepare\_armor(cat, canary), give\_magnifier(cricket, canary)), Not(remove\_piece(canary, aardvark)))\\
\# Rule4: If the cat has a name whose first letter is the same as the first letter of the puffin's name, then the cat does not prepare armor for the canary.\\
Rule4 = Implies(has\_same\_first\_letter\_name(cat, puffin), Not(prepare\_armor(cat, canary)))\\
\# Rule5: The canary removes one of the pieces of the aardvark whenever at least one animal proceeds to the spot right after the sun bear.\\
Rule5 = Implies(Exists([x], proceed\_to\_spot\_right\_after(x, sun\_bear)), remove\_piece(canary, aardvark))\\
\# Rule6: Regarding the cat, if it has something to sit on, then we can conclude that it does not prepare armor for the canary.\\
Rule6 = Implies(has\_something\_to\_sit\_on(cat), Not(prepare\_armor(cat, canary))) \\  
\# The current state of the game is as follows. The cat is named Meadow. The cat published a high-quality paper. The gecko winks at the eagle. The puffin is named Peddi.\\
\# The cat is named Meadow. The puffin is named Peddi. The first letter of Meadow is M. The first letter of the Peddi is P. So the cat does not have the same first letter name as the puffin.\\
has\_same\_first\_letter\_name(cat, puffin) == False\\
\# The cat published a high-quality paper.\\
has\_high\_quality\_paper(cat) == True\\
\# The gecko winks at the eagle.\\
winks\_at(gecko, eagle) == True\\
\# preferences. Rule3 is preferred over Rule5. Rule4 is preferred over Rule1. Rule6 is preferred over Rule1.\\
soft\_rules = [Rule5, Rule1, Rule1]\\
\# Rule3 is preferred over Rule5. So Rule5 is suppressed by the precondition of Rule3.\\
Rule5 = Or(And(prepare\_armor(cat, canary), give\_magnifier(cricket, canary)), Rule5)\\
\# Rule4 is preferred over Rule1. So Rule1 is suppressed by the precondition of Rule4.\\
Rule1 = Or(has\_same\_first\_letter\_name(cat, puffin), Rule1)\\
\# Rule6 is preferred over Rule1. So Rule1 is suppressed by the precondition of Rule6.\\
Rule1 = Or(has\_something\_to\_sit\_on(cat), Rule1)\\
\# question: does the canary remove from the board one of the pieces of the aardvark?\\
return remove\_piece(canary, aardvark)\\
\end{boxK}

\subsection{GPT4o and Cohere command-r-plus Prompts} \label{GPT4O Prompts}
The prompts require some adjustments for GPT-4O and Cohere, as both models tend to produce complete executable code rather than adhering to the provided example. For instance, GPT-4O will define "s.solver()" and create the decision rule for Z3, instead of generating translations as specified in the prompt. Here we provide an overview of what is changed in the prompt. 
\paragraph{ProofWriter GPT4O Prompts for Z3 Solver} One-shot demonstration 
\begin{boxK}\label{Z3 ProofWriter Prompts GPT4o}
The grammar of the first-order logic formula is defined as follows:\\
1) logical conjunction of expr1 and expr2: And(expr1, expr2)\\
2) logical disjunction of expr1 and expr2: Or(expr1, expr2)\\
3) logical exclusive disjunction of expr1 and expr2: Xor(expr1, expr2)\\
4) logical negation of expr1: Not(expr1)\\
5) expr1 implies expr2: Implies(expr1, expr2)\\
6) expr1 if and only if expr2: expr1 == expr2\\
7) logical universal quantification: ForAll()\\
8) logical existential quantification: Exists()\\
Given a problem description and a question. The task is to parse the {[}Problem{]} and the {[}Question{]} into Python Z3 solver. You are meant to follow the example format and do not provide any further explanations. Keep all the \# signs as symbols and do not interpret them as markdown marker.\\
------\\
{[}Problem{]}:\\
Anne is quiet. 
Erin is furry. 
Erin is green. 
Fiona is furry. 
Fiona is quiet. 
Fiona is red. 
Fiona is rough. 
Fiona is white. 
Harry is furry. 
Harry is quiet. 
Harry is white. 
Young people are furry. 
If Anne is quiet then Anne is red. 
Young, green people are rough. 
If someone is green then they are white. 
If someone is furry and quiet then they are white. 
If someone is young and white then they are rough. 
All red people are young.

{[}Question{]}:\\
Based on the above information, is the following statement true, false, or unknown? Anne is white.
\#\#\#\#

{[}Problem Parse Output{]}:\\
\# Anne is quiet.

quiet(Anne)

\# Erin is furry.

furry(Erin)

\# Erin is green.

green(Erin)

\# Fiona is furry.

furry(Fiona)

\# Fiona is quiet.

quiet(Fiona)

\# Fiona is red.

red(Fiona)

\# Fiona is rough.

rough(Fiona)

\# Fiona is white.

white(Fiona)

\# Harry is furry.

furry(Harry)

\# Harry is quiet.

quiet(Harry)

\# Harry is white.

white(Harry)

\# Young people are furry.

ForAll([x], Implies(young(x), furry(x)))

\# If Anne is quiet then Anne is red.

Implies(quiet(Anne), red(Anne))

\# Young, green people are rough.

ForAll([x], Implies(And(young(x), green(x)), rough(x)))

\# If someone is green then they are white.

ForAll([x], Implies(green(x), white(x)))

\# If someone is furry and quiet then they are white.

ForAll([x], Implies(And(furry(x), quiet(x)), white(x)))

\# If someone is young and white then they are rough.

ForAll([x], Implies(And(young(x), white(x)), rough(x)))

\# All red people are young.

ForAll([x], Implies(red(x), young(x)))\\
{[}Question Parse Output{]}:\\
\# Question: the following statement true, false, or unknown? Anne is white.

return white(Anne)
\end{boxK}
\paragraph{ProofWriter Cohere Prompts for Z3 Solver} One-shot demonstration \\
For the Z3 solver, the Cohere prompt was slightly adjusted because produces translation not aligned with the given example. 
\begin{boxK}
The grammar of the first-order logic formula is defined as follows:\\
1) logical conjunction of expr1 and expr2: And(expr1, expr2)\\
2) logical disjunction of expr1 and expr2: Or(expr1, expr2)\\
3) logical exclusive disjunction of expr1 and expr2: Xor(expr1, expr2)\\
4) logical negation of expr1: Not(expr1)\\
5) expr1 implies expr2: Implies(expr1, expr2)\\
6) expr1 if and only if expr2: expr1 == expr2\\
7) logical universal quantification: ForAll()\\
8) logical existential quantification: Exists()\\
Given a problem description and a question. The task is to parse the {[}Problem{]} and the {[}Question{]} into Python Z3 solver. You are meant to follow the example format and do not provide any further explanations.  \textcolor{red}{Follow the format given and do not define "s" and "s.solver" for the Z3 solver.} Keep all the \# signs as symbols and do not interpret them as markdown marker.\\
------\\
{[}Problem{]}:\\
Anne is quiet. \\
.......
\end{boxK}

\paragraph{ProofWriter GPT4o and Cohere Prompts for Prover9} One shot demonstration for LLM

\begin{boxK}
 The grammar of the first-order logic formula is defined as follows:
\begin{enumerate}
    \item Logical conjunction of $\text{expr1}$ and $\text{expr2}$: $\text{expr1} \land \text{expr2}$
    \item Logical disjunction of $\text{expr1}$ and $\text{expr2}$: $\text{expr1} \lor \text{expr2}$
    \item Logical exclusive disjunction of $\text{expr1}$ and $\text{expr2}$: $\text{expr1} \oplus \text{expr2}$
    \item Logical negation of $\text{expr1}$: $\neg \text{expr1}$
    \item $\text{expr1}$ implies $\text{expr2}$: $\text{expr1} \rightarrow \text{expr2}$
    \item $\text{expr1}$ if and only if $\text{expr2}$: $\text{expr1} \leftrightarrow \text{expr2}$
    \item Logical universal quantification: $\forall x$
    \item Logical existential quantification: $\exists x$
\end{enumerate}
Given a problem description and a question. The task is to parse the {[}Problem{]} and the {[}Question{]} into Prover9 solver. You are meant to follow the example format and do not provide any further explanations. Keep all the ::: signs as symbols and do not interpret them as markdown marker.\\
{[}Problem{]}: \\
Anne is quiet. 
Erin is furry. 
Erin is green. 
Fiona is furry. 
Fiona is quiet. 
Fiona is red. 
Fiona is rough. 
Fiona is white. 
Harry is furry. 
Harry is quiet. 
Harry is white. 
Young people are furry. 
If Anne is quiet then Anne is red. 
Young, green people are rough. 
If someone is green then they are white. 
If someone is furry and quiet then they are white. 
If someone is young and white then they are rough. 
All red people are young.

{[}Question{]}:\\
Based on the above information, is the following statement true, false, or unknown? Anne is white.
\#\#\#\#

{[}Problem Parse Output{]}:\\
\text{Predicates} \\
\text{quiet(x)} :::  \text{x is quiet.}\\
\text{furry(x)} ::: \text{x is furry.} \\
\text{green(x)} ::: \text{x is green.}\\
\text{red(x)} ::: \text{x is red.}\\
\text{rough(x)} ::: \text{x is rough.}\\
\text{white(x)} ::: \text{x is white.}\\
\text{young(x)} ::: \text{x is young}\\
\text{Premises} \\
\text{quiet(Anne)} :::  \text{Anne is quiet.}\\
\text{furry(Erin)} ::: \text{Erin is furry.} \\
\text{green(Erin)} ::: \text{Erin is green.}\\
\text{furry(Fiona)} ::: \text{Fiona is furry.} \\
\text{quiet(Fiona)} ::: \text{Fiona is quiet.} \\
\text{red(Fiona)} ::: \text{Fiona is red.}\\
\text{rough(Fiona)} ::: \text{Fiona is rough.}\\
\text{white(Fiona)} ::: \text{Fiona is white.}\\
\text{furry(Harry)} ::: \text{Harry is furry.}\\
\text{quiet(Harry)} ::: \text{Harry is quiet.}\\
\text{white(Harry)} ::: \text{Harry is white.}\\
$\forall x (\text{young}(x) \rightarrow \text{furry}(x)) \,:::\, \text{Young people are furry.}$ \\
$(\text{quiet(Anne)} \rightarrow \text{red(Anne)}) ::: \text{If Anne is quiet then Anne is red.}$\\
$\forall x (\text{young}(x) \land \text{green}(x) \rightarrow \text{rough}(x))$ ::: \text{Young, green people are rough.} \\
$\forall x (\text{green}(x) \rightarrow \text{white}(x)) ::: \text{If someone is green then they are white.} $\\
$\forall x ((\text{furry}(x) \land \text{quiet}(x)) \rightarrow \text{white}(x)) ::: \text{If someone is furry and quiet then they are white.}$ \\
$\forall x ((\text{young}(x) \land \text{white}(x)) \rightarrow \text{rough}(x)) ::: \text{If someone is young and white then they are rough.}$\\
$\forall x (\text{red}(x) \rightarrow \text{young}(x)) ::: \text{All red people are young.}$\\
{[}Question Parse Output{]}:\\
\text{Conclusion:}\\
\text{white(Anne)}

\end{boxK}

\paragraph{ProofWriter GPT4o and Cohere Prompts for Pyke Solver} One-shot demonstration

\begin{boxK}
The grammar of the first-order logic formula is defined as follows:\\
1) logical conjunction of expr1 and expr2: expr1 \&\& expr2\\
2) logical negation of expr1: expr1(\$x, False), as example if "Anne is not quiet", the term would be "Quiet(Anne, False)"\\
3) expr1 implies expr2: expr1 >>> expr2\\
Given a problem description and a question. The task is to parse the {[}Problem{]} and the {[}Question{]} into Pyke solver. You are meant to follow the example format and do not provide any further explanations. Keep all the ::: signs as symbols and do not interpret them as markdown marker.\\
{[}Problem{]}: \\
Anne is quiet. 
Erin is furry. 
Erin is green. 
Fiona is furry. 
Fiona is quiet. 
Fiona is red. 
Fiona is rough. 
Fiona is white. 
Harry is furry. 
Harry is quiet. 
Harry is white. 
Young people are furry. 
If Anne is quiet then Anne is red. 
Young, green people are rough. 
If someone is green then they are white. 
If someone is furry and quiet then they are white. 
If someone is young and white then they are rough. 
All red people are young.

{[}Question{]}:\\
Based on the above information, is the following statement true, false, or unknown? Anne is white.\\
\#\#\#\#\\
{[}Problem Parse Output{]}:\\
\text{Predicates:} \\
\text{quiet(\$x, bool)} :::  \text{x is quiet.}\\
\text{furry(\$x, bool)} ::: \text{x is furry.} \\
\text{green(\$x, bool)} ::: \text{x is green.}\\
\text{red(\$x, bool)} ::: \text{x is red.}\\
\text{rough(\$x, bool)} ::: \text{x is rough.}\\
\text{white(\$x, bool)} ::: \text{x is white.}\\
\text{young(\$x, bool)} ::: \text{x is young}\\
\text{Facts:} \\
\text{quiet(Anne, True)} :::  \text{Anne is quiet.}\\
\text{furry(Erin, True)} ::: \text{Erin is furry.} \\
\text{green(Erin, True)} ::: \text{Erin is green.}\\
\text{furry(Fiona, True)} ::: \text{Fiona is furry.} \\
\text{quiet(Fiona, True)} ::: \text{Fiona is quiet.} \\
\text{red(Fiona, True)} ::: \text{Fiona is red.}\\
\text{rough(Fiona, True)} ::: \text{Fiona is rough.}\\
\text{white(Fiona, True)} ::: \text{Fiona is white.}\\
\text{furry(Harry, True)} ::: \text{Harry is furry.}\\
\text{quiet(Harry, True)} ::: \text{Harry is quiet.}\\
\text{white(Harry, True)} ::: \text{Harry is white.}\\
\text{young}(\$x, True) $>>>$ \text{furry}(\$x, True)) \,:::\, \text{Young people are furry.} \\
\text{quiet(Anne, True)} $>>>$ \text{red(Anne, True)}) ::: \text{If Anne is quiet then Anne is red.}\\
\text{young}(\$x, True) \&\& \text{green}(\$x, True) $>>>$ \text{rough}(\$x, True) ::: \text{Young, green people are rough.} \\
\text{green}(\$x, True) $>>>$ \text{white}(\$x, True) ::: \text{If someone is green then they are white.} \\
\text{furry}(\$x, True) \&\& \text{quiet}(\$x, True) $>>>$ \text{white}(\$x, True) \\::: \text{If someone is furry and quiet then they are white.} \\
\text{young}(\$x, True) \&\& \text{white}(\$x, True) $>>>$ \text{rough}(\$x, True)\\ ::: \text{If someone is young and white then they are rough.}\\
\text{red}(\$x, True)  $>>>$ \text{young}(\$x, True) ::: \text{All red people are young.} \\
{[}Question Parse Output{]}:\\
\text{Query:}\\
\text{white(Anne)}
\end{boxK}




\begin{table*}[b]
\centering
\begin{tabular}{llll}
\hline
\textbf{Dataset}   & \textbf{Z3}                    & \textbf{Prover9}               & \textbf{Pyke} \\ \cline{2-4} 
                   & Avg\_Acc                       & Avg\_Acc                       & Avg\_Acc      \\ \hline
ProofWriter D5 OWA & {\color[HTML]{FE0000} 85.75\%} & 75.00\%                        & 56.63\%       \\ 
ProofWriter D3 OWA & {\color[HTML]{FE0000} 83.04\%} & 75.37\%                        & 52.37\%       \\ 
ProofWriter D2 OWA & {\color[HTML]{FE0000} 82.50\%} & 76.00\%                        & 56.50\%       \\ 
ProofWriter D5 CWA & {\color[HTML]{FE0000} 87.50\%} & 78.25\%                        & 60.25\%       \\ 
ProofWriter D3 CWA & {\color[HTML]{FE0000} 89.25\%} & 76.13\%                        & 45.63\%       \\ 
ProofWriter D2 CWA & {\color[HTML]{FE0000} 89.63\%} & 77.12\%                        & 54.75\%       \\ 
PrOntoQA           & {\color[HTML]{FE0000} 94.12\%} & 76.87\%                        & 91.12\%       \\ 
FOLIO (1 Shot)     & 26.25\%                        & {\color[HTML]{FE0000} 43.78\%} &     \ding{55}          \\ 
FOLIO (2 Shot)     & 34.36\%                        & {\color[HTML]{FE0000} 41.60\%} &     \ding{55}          \\ 
FOLIO (4 Shot)     & 36.87\%                        & {\color[HTML]{FE0000} 49.13\%} &      \ding{55}         \\ \hline
\end{tabular}
\caption{Average accuracy of Experiment done with GPT-4o, GPT-3.5-turbo, Gemini-1.0-pro and command-r-plus on all datasets. We present the percentage of the overall average accuracy of tools (Avg\_Acc). The shots represent the number of shots used in the prompt.\ding{55}: the tool was unable to solve this dataset. The numbers highlighted in red color represent the highest accuracy between the 3 chosen tools.}
\label{Average Main table}
\end{table*}



\end{document}